\documentclass[runningheads]{llncs}
\usepackage[T1]{fontenc}
\usepackage{graphicx}
\usepackage[numbers]{natbib}
\begin{document}
\title{Advancing Student Writing Through Automated Syntax Feedback}
%

%
\author{Kamyar Zeinalipour\inst{1} \and
Mehak Mehak\inst{1} \and
Fatemeh Parsamotamed\inst{1}\and
Marco Maggini\inst{1}\and
Marco Gori\inst{1}}

\authorrunning{K. Zeinalipour et al.}
%

\institute{University of Siena, Siena, Italy\\
\email{kamyar.zeinalipour2@unisi.it}}
\maketitle              









\begin{abstract}
This study underscores the pivotal role of syntax feedback in augmenting the syntactic proficiency of students. Recognizing the challenges faced by learners in mastering syntactic nuances, we introduce a specialized dataset named \texttt{Essay-Syntax-Instruct} designed to enhance the understanding and application of English syntax among these students. Leveraging the capabilities of Large Language Models (LLMs) such as \textit{GPT3.5-Turbo}, \textit{Llama-2-7b-chat-hf}, \textit{Llama-2-13b-chat-hf}, and \textit{Mistral-7B-Instruct-v0.2}, this work embarks on a comprehensive fine-tuning process tailored to the syntax improvement task. Through meticulous evaluation, we demonstrate that the fine-tuned LLMs exhibit a marked improvement in addressing syntax-related challenges, thereby serving as a potent tool for students to identify and rectify their syntactic errors. The findings not only highlight the effectiveness of the proposed dataset in elevating the performance of LLMs for syntax enhancement but also illuminate a promising path for utilizing advanced language models to support language acquisition efforts. This research contributes to the broader field of language learning technology by showcasing the potential of LLMs in facilitating the linguistic development of Students. \end{abstract}


\section{Introduction}

Mastering syntax, which is essentially the set of rules governing how sentences are structured in any given language, stands as a major obstacle in the process of learning English. This challenge is formidable not only for students who are non-native speakers but also for those who have grown up speaking the language. English syntax is notorious for its complexity, with a wide array of rules and numerous exceptions to those rules, making the path to achieving fluency and proficiency a daunting one. Traditional methods of teaching, while valuable in their own right, often do not fully address the unique needs of each learner, leaving gaps in understanding and application of syntactic rules. In the face of these challenges, the burgeoning field of artificial intelligence (AI) and natural language processing (NLP) presents new opportunities for language education. Specifically, the development of Large Language Models (LLMs) has introduced the potential for more personalized, adaptable, and effective learning tools. These models, capable of generating text that closely mimics human writing, represent a significant breakthrough in approaches to language learning, especially in the intricate area of syntax.\\
In this study, we present a framework that capitalizes on the most recent Large Language Model (LLM) technology, with a particular focus on models such as \textit{GPT3.5-Turbo}, \textit{Llama-2-7b-chat-hf}, L\textit{Llama-2-13b-chat-hf}, and \textit{Mistral-7B-Instruct-v0.2}. Our objective is to significantly improve English learners' syntactic skills by tackling the challenges they encounter in mastering English syntax. Understanding and applying the complex rules of English syntax is a daunting task for learners, often impeding their progress. The traditional methods of teaching English syntax, despite their benefits, may not fully cater to the varied needs of learners, highlighting the demand for a more personalized and adaptable learning strategy.\\
Our research leverages these state-of-the-art LLMs to address the intricate challenges associated with learning English syntax. We have developed a specialized dataset aimed at improving both the understanding and application of syntactic rules. This endeavor also involves fine-tuning these LLMs to better serve educational tasks focusing on syntax, thereby making a significant leap in creating tailor-made language learning aids.\\
We also invite fellow researchers to delve into this unique dataset, thereby setting a new benchmark in language learning technology research. By applying our dataset, these LLMs undergo a meticulous fine-tuning process, specifically designed to bolster syntactic proficiency. This process markedly enhances the models' ability to offer personalized and effective learning experiences. Through our work, we demonstrate a substantial advancement in the use of LLMs for educational purposes, particularly in the domain of syntax learning, offering learners a more intuitive and adaptable means to master the complexities of English syntax.\\
By detailing the creation of our specialized dataset and the further development of tailor-made LLM-based models, the paper extends its contributions beyond a simple technical discourse. It sets a solid groundwork for additional research, offering a comprehensive framework for assessing the effectiveness of models before and after fine-tuning, using our unique dataset as a reference point. Our holistic approach demonstrates not only the tangible benefits of our intervention in improving the feedback capabilities of LLMs but also provides the academic and educational spheres with a valuable toolkit for addressing the persistent challenges of mastering syntax in language learning. Through these efforts, the study introduces new resources and fine-tuned LLMs to the field of linguistics and paves the way for future exploration into leveraging technology to break down linguistic barriers, showcasing the profound potential of AI and LLMs in transforming language education.\\
We made the \texttt{Essay-Syntax-Instruct} dataset \footnote{https://huggingface.co/datasets/Kamyar-zeinalipour/Essay-Syntax-Instruct} and all models \footnote{https://huggingface.co/Kamyar-zeinalipour/Mistral7B-Syntax-Instruct\\ https://huggingface.co/Kamyar-zeinalipour/Llama2-13B-Syntax-Instruct\\ https://huggingface.co/Kamyar-zeinalipour/Llama2-7B-Syntax-Instruct} publicly available to contribute to the open-source community, supporting other researchers in using, developing, and extending this type of study. Additionally, it can potentially be used by schools and students to significantly improve English learners' syntactic skills by addressing the challenges they face in mastering English syntax.\\
The structure of this paper is outlined as follows: section~\ref{sec:relatedworks} reviews the relevant literature in the field. Our approach is detailed in section~\ref{sec:methodology}, followed by a comprehensive analysis of the properties of the generated dataset in section~\ref{sec:dataset}. The results of our experiments are discussed extensively in section~\ref{sec:experiments}, we conclude our findings in section~\ref{sec:conclusions}

\section{Related Works}~\label{sec:relatedworks}

The integration of technology in educational assessment has led to significant advancements in Automated Essay Scoring (AES) and Essay Feedback Generation. AES systems utilize machine learning algorithms to evaluate and score written essays with efficiency and consistency, mirroring the grading process traditionally conducted by human educators. Complementing AES, Essay Feedback Generation focuses on the qualitative aspect of writing improvement. Together, these technologies represent pivotal tools in the digital education landscape, facilitating scalable assessment and personalized learning experiences. This section delves into the nuances of AES and Essay Feedback Generation, laying the groundwork for a detailed discussion of each area's contributions, challenges, and potential synergies with our proposed Automatic Syntax Feedback System.

\subsection{Automated Essay Scoring}

Automated Essay Scoring (AES) represents a vital interdisciplinary field that merges natural language processing (NLP) with educational technology, driving toward the automation of student essay evaluations. Traditional approaches in this domain have followed a dual-step methodology, initiating with the extraction of a broad range of features from texts — spanning statistical to linguistic indicators \cite{miltsakaki2004evaluation,ridley2020prompt} — and followed by the deployment of regression or classification models for predicting essay scores \cite{sultan2016fast,mathias2018thank,salim2019automated}. The advent and subsequent rise of deep learning have facilitated the inclusion of more sophisticated techniques such as convolutional neural networks (CNNs), long short-term memory networks (LSTMs), and attention-based models, significantly enhancing the precision of scoring algorithms \cite{dong2016automatic,taghipour2016neural,riordan2017investigating}.

The AES field has seen further refinement with the integration of pre-trained language learning models (LLMs), leveraging their advanced capabilities to push the boundaries of essay scoring. This integration has been marked by the employment of models like BERT for nuanced essay evaluation strategies \cite{rodriguez2019language,lun2020multiple}, and the innovation of enhancing BERT's performance through specialized fine-tuning approaches \cite{wang2022use,yang2020enhancing}. Despite these advancements, efforts to align the latest LLMs, such as GPT, with AES tasks have encountered mixed outcomes. These models, when tuned to specific scoring rubrics, have not consistently surpassed the benchmark set by traditional methodologies or human evaluators in providing accurate scoring \cite{mizumoto2023exploring,han2023fabric,yancey2023rating}.\\
Leveraging insights from AES developments, our work introduces the "Automatic Syntax Feedback System," a novel solution addressing key challenges within the AES field, such as the lack of expert-annotated, rubric-based datasets and the intricacies of essay scoring. This system emphasizes the syntactic improvement of student essays, offering prompt, constructive feedback, and thus, bridges a crucial gap in AES research. By harnessing deep learning and the precision of modern LLMs, our approach not only supports accurate scoring but also plays a vital role in enhancing students' writing proficiency, marking a significant advancement in AES applications.

\subsection{Essay Feedback Generation}

The exploration of Language Learning Models (LLMs) in education has recently pivoted towards their potential to revolutionize feedback generation, providing real-time and personalized insights for learners \cite{yan2024practical,kasneci2023chatgpt,xiao2024automation}. Despite the burgeoning interest, there remains a gap in thoroughly investigating how LLMs can be intricately applied to create detailed educational feedback mechanisms. \cite{peng2023check} made strides in illustrating how LLMs, when equipped with specific and rich information - termed as "golden knowledge", exhibit significant improvements in areas such as task-oriented dialogue and open-domain question answering. This underscores the value of integrating precise, domain-specific knowledge, like rubric explanations and accurate essay evaluations, into LLMs to enhance their capability to generate more constructive and targeted feedback on student essays.\\
In this section, we integrate these findings to underscore our research on the "Automatic Syntax Feedback System." By emphasizing the syntactic elements of essays, our system notably leverages the advancements in feedback generation propelled by the enhanced capabilities of modern LLMs. Recognizing the crucial role of detailed, golden knowledge in empowering LLMs, our approach enriches these models with a comprehensive understanding of English syntax and writing pedagogies. This strategic incorporation aims to fine-tune the feedback provided, making it not only immediate but also pedagogically valuable, thereby advancing the realms of automated essay scoring and educational support systems.

\section{Methodology}\label{sec:methodology}
\begin{figure*}[ht]
    \centering 
    \includegraphics[width=0.9\textwidth]{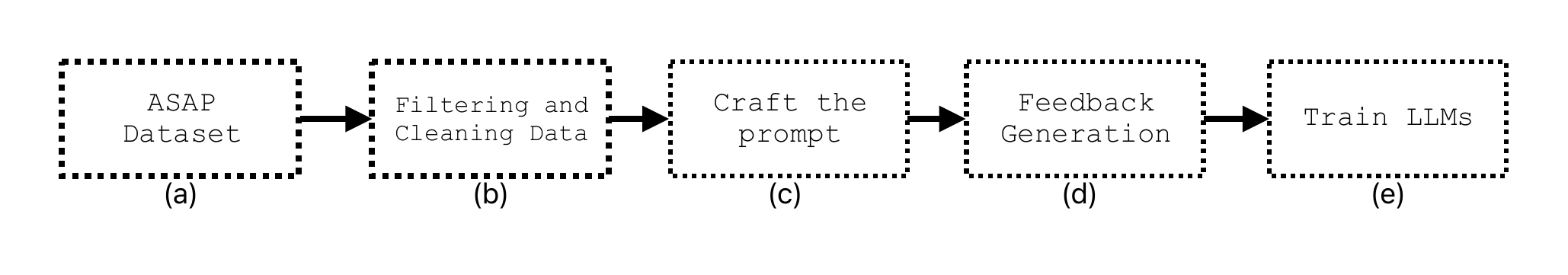}
    \caption{This figure illustrates the pipeline employed in this study: (a) Acquiring essays authored by humans from the ASAP dataset. (b) Data refinement and filtering to enhance data quality, by the essay placeholder replacement, eliminating excessively short or overly detailed essays. (c) Design of the prompt for generating syntax feedback based on input essay. (d) Exploit of \textit{GPT3.5-Turbo} to generate syntax feedback from the collected data and defined prompts. (e) Fine-tune Large Language Models (LLMs) to enhance their ability to provide syntax feedback. }
    \label{fig:fig1}
\end{figure*}

In this study, we developed an Automatic Syntax Feedback Generation system. As a foundational step, we compiled a dataset by utilizing the widely recognized ASAP \footnote{https://www.kaggle.com/c/asap-aes} dataset, which is frequently employed in Automated Essay Scoring (AES) tasks. This foundational dataset comprises essays from students in Grades 7-10, across eight distinct prompts. While the original purpose of the ASAP dataset was for scoring essays, our objective diverged as we aimed to generate automatic syntax feedback from these essays. To adapt this dataset to our needs, we focused on employing essays written by students, alongside leveraging Large Language Models (LLMs) and prompt engineering techniques to produce syntax feedback. Subsequently, the generated feedback was subjected to evaluation by linguistic experts. Based on their insights, we refined our dataset for the final phase of fine-tuning LLMs, including models such as \textit{Llama-2-7b-chat-hf}, \textit{Mistral-7B-Instruct-v0.2}, and \textit{Llama-2-13b-chat-hf}, to specialize in this task.\\
Figure \ref{fig:fig1} illustrates the overarching methodology pipeline of this study. Following this, we delve into a discussion centered on the dataset generation process and the specific models we employed to accomplish this task.

\subsection{\texttt{Essay-Syntax-Instruct} Dataset}\label{sec:dataset}
As we discussed earlier we used the ASAP dataset in the study presented in this paper, a significant and unique data set originating from a competitive environment was employed to scrutinize the automated scoring engine's capability in evaluating written essays. The data set in question encompasses responses to eight distinct essay prompts, resulting in a collection of essays varied in both theme and response length, which runs from an average of 150 to 550 words per essay. These essays were produced by students across a range of academic grades, specifically from Grade 7 to Grade 10. To ensure the credibility and reliability of the scoring, each essay underwent a rigorous evaluation process, being graded by hand and subject to a double-scoring protocol. Reflecting the diverse nature of the essay prompts, the responses exhibit a wide range of characteristics, from dependence on source material to standalone narratives, thus providing a comprehensive test bed for the scoring engines under examination.\\
\paragraph{Place Holder Replacement}
In this study, we initially focused on utilizing a dataset intended for Automated Essay Scoring for our Automatic Syntax Feedback System. This involved leveraging essays written by students to generate feedback using Large Language Models (LLMs). A significant challenge we encountered was related to the essays in the ASAP dataset, specifically due to the comprehensive anonymization effort undertaken to protect student privacy. This process involved the use of Stanford's Natural Language Processing group's Named Entity Recognizer (NER) tool, among other techniques, to replace identifiable information (such as names, organizations, locations, dates, times, money, and percent values) with anonymized placeholders (e.g., "@PERSON1"). Although necessary for privacy, this introduced inefficiencies when the LLMs interpreted these placeholders as errors, impacting the quality of the feedback generated.\\
\begin{figure}[ht]
  \centering
  \includegraphics[width=0.8\textwidth]{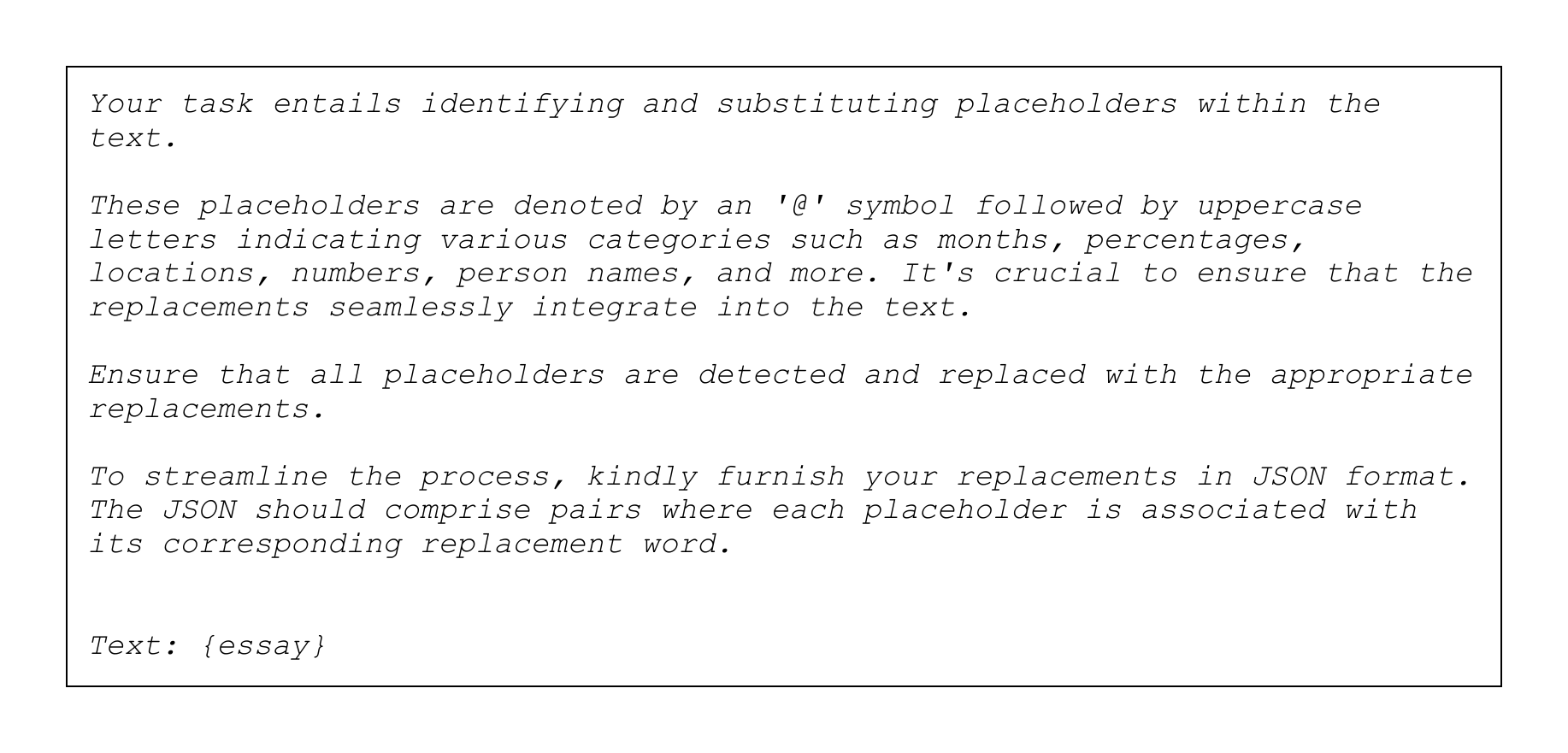}
  \caption{Place Holder Replacement Prompt.}
  \label{fig:fig2}
\end{figure}
To address this issue, we employed \textit{GPT3.5-Turbo}, prompting the model to replace placeholders with contextually coherent alternatives that blend seamlessly with the rest of the essay content. This approach significantly improved the LLMs' ability to provide relevant and accurate feedback. We illustrate the prompt used for this task in Figure \ref{fig:fig2}, showcasing our strategy to overcome the anonymization challenge while maintaining the ethical use of \texttt{Essay-Syntax-Instruct} dataset.

\begin{figure*}[ht]
    \centering 
    \includegraphics[width=0.9\textwidth]{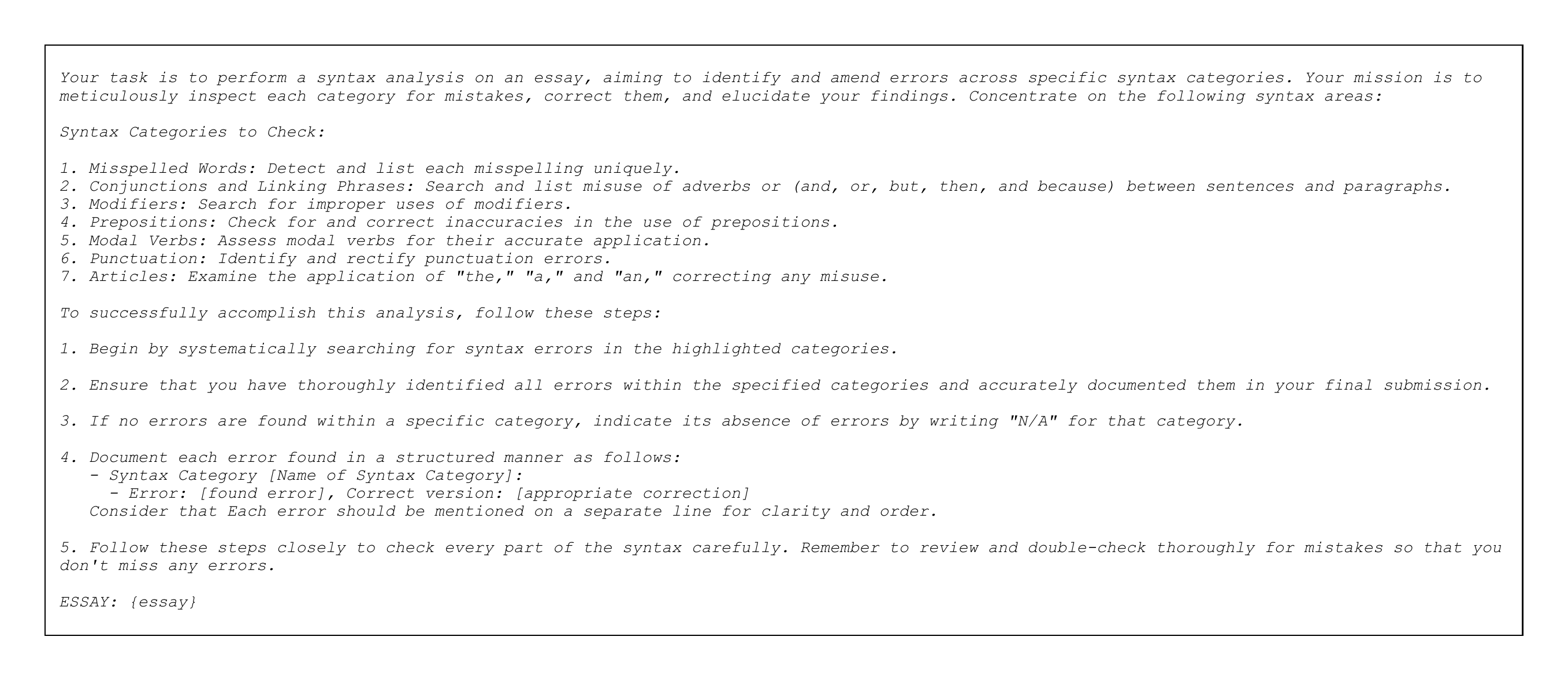}
    \caption{Syntax Feedback Generation Prompt.}
    \label{fig:fig3}
\end{figure*}

\paragraph{Syntax Feedback Generation}
After substituting placeholders with suitable replacements, our next objective was to generate syntax feedback for each essay. 
To achieve this, we utilized \textit{GPT3.5-Turbo}, specifically \textit{GPT3.5-Turbo-0125}, with a temperature setting of 0.3 with a tailored prompt designed to elicit syntax feedback relevant to each essay. This prompt sought to examine various syntax components including misspelled words, conjunctions and linking phrases, modifiers, prepositions, modal verbs, punctuation, and articles. These categories were chosen because they cover the majority of syntax errors found in student essays, allowing for comprehensive and effective feedback that targets the most common issues. Categories of syntax feedback given by real instructors also align with the GPT-generated feedback, except for the grammatical errors category. During our dataset preparation, we noticed that GPT-generated feedback effectively corrected all the errors within a single grammar category. This redundancy led us to omit that one grammar error category, as the other syntax categories were already specified to address those errors. Figure \ref{fig:fig3} illustrates the prompt utilized for this purpose.

\begin{figure*}[ht]
    \centering 
    \includegraphics[width=0.9\textwidth]{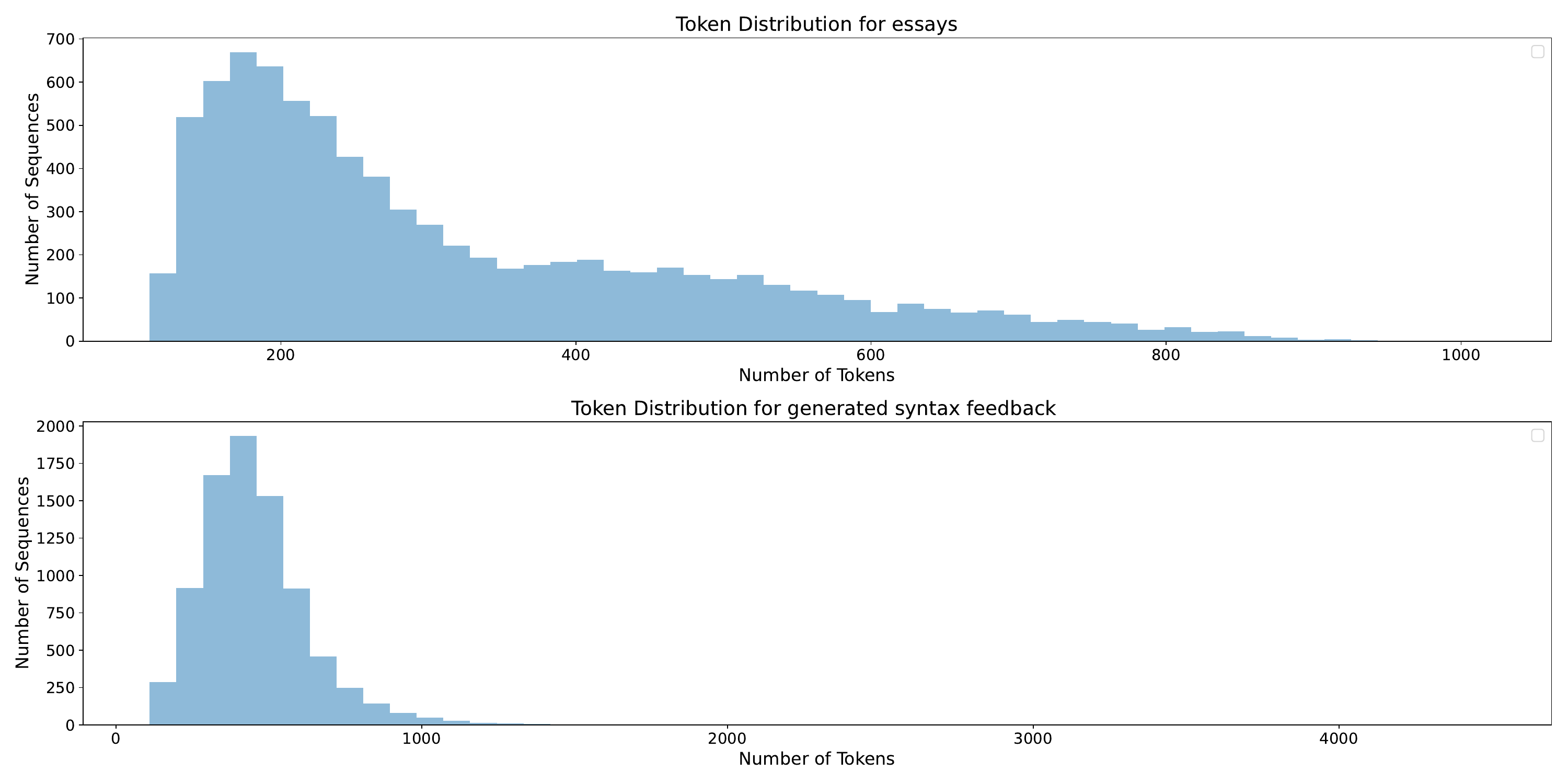}
    \caption{Token Distribution of essays and generated syntax feedback }
    \label{fig:fig4}
\end{figure*}

Subsequently, we obtained a dataset comprising essays—devoid of placeholders—written by students, alongside the corresponding feedback generated by the model across the aforementioned syntax categories. To ensure conciseness and relevance, we refined the dataset by excluding essays that were excessively long or short, focusing on those ranging from 100 to 700 words. Following data filtering, we are left with a dataset comprising 8,320 examples of essay and syntax feedback. Additionally, \ref{fig:fig4} displays the distribution of tokens within the generated feedback and the essays. For this analysis, we utilized the \textit{Llama-2} tokenizer.\\
\paragraph{Data Set Quality Evaluation}

\begin{figure}[ht]
  \centering
  \includegraphics[width=0.9\textwidth]{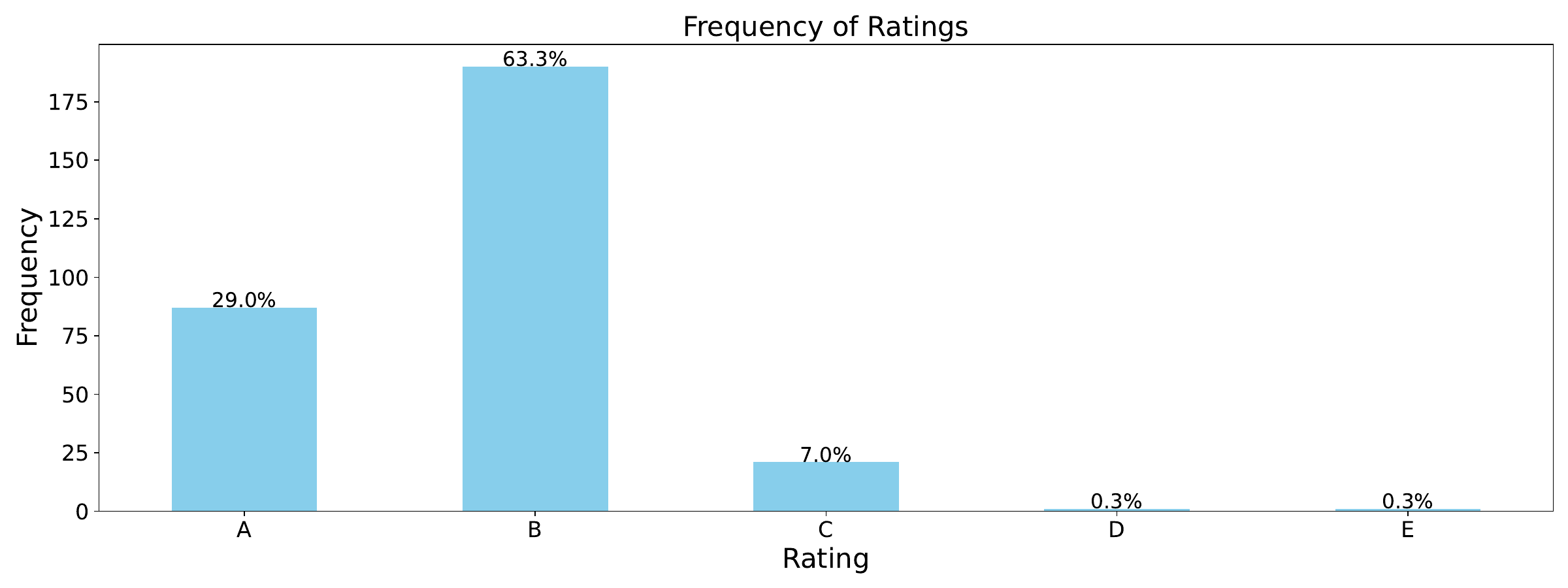}
  \caption{Assessment results of human evaluations for the syntax feedback produced by \texttt{GPT3.5-Turbo}.}
  \label{fig:fig5}
\end{figure}

Due to the lack of a reference corpus specifically for Essay Syntax Feedback, establishing a baseline for comparison of generated Syntax Feedback is challenging. This absence makes it difficult to apply metrics like ROGUE scores to evaluate the effectiveness of educational feedback generation. Consequently, we opted for a qualitative evaluation approach via human assessment. Specifically, we randomly selected a sample of clues and had them evaluated by a panel of experts in linguistics. This method mirrors the approach taken by \cite{wang2022self}, utilizing a five-tier rating system guided by predetermined criteria:
\begin{itemize}
    \item \texttt{RATING-A}: Feedback is exceptional, systematically identifying every error in the text and clearly mentioning these alongside their accurate corrections. It also enriches the review with substantial suggestions or revised sentences and exhibits a profound comprehension of the content's nuances. The structured approach of this feedback, which methodically points out errors and provides accurate corrections, enhances the overall quality of the review and demonstrates a thorough understanding of the text.

    \item \texttt{RATING-B}: Feedback is robust, pinpointing most errors and correctly articulating them with their respective corrections. It provides constructive criticism but might overlook minor issues. Nonetheless, it demonstrates a high level of engagement with the content’s essence. The structured format of this feedback, showcasing a detailed analysis of errors and their corrections, adds value to the review and reflects a strong engagement with the essence of the text.

    \item \texttt{RATING-C}: Feedback adequately identifies and reports on a considerable amount of errors, offering appropriate corrections. While the suggestions are helpful, they could be more specifically tailored to enhance the text's overall quality and effectiveness. The structure of this feedback, though adequate in identifying errors and providing corrections, could benefit from a more tailored approach to offer specific enhancements for the text's improvement.
    \item \texttt{RATING-D}: Feedback catches some errors and presents corrections but does so inconsistently. The feedback contains basic suggestions, lacking the depth and detail necessary to facilitate significant improvement in the text. The lack of consistency in the structure of this feedback affects its overall effectiveness in guiding improvements and enhancing the text's quality.
    \item \texttt{RATING-E}: Feedback is deficient, with little to no identification of the main errors, offering few if any corrections. It fails to provide informative guidance and does not contribute meaningfully to the betterment of the text’s accuracy or readability. The absence of a structured approach in this feedback results in a lack of clarity and guidance, hindering its ability to improve the text effectively.
\end{itemize}
During this human evaluation process, we analyzed 300 examples of essay feedback. The outcomes of this evaluation are depicted in Figure~\ref{fig:fig5}. In the assessment conducted via human evaluation, the feedback generated by GPT-3.5 did not attain 100\% accuracy. As depicted in Figure~\ref{fig:fig5}, Rating A is recorded at 29\%, while Rating B exhibits a remarkably higher percentage of 63.3\%, signifying an exceptional level of quality deemed satisfactory in approximately 80-90\% of instances, Rating B notably stands out as exceptionally good.

The results of the human evaluation indicate that the quality of the final \texttt{Essay-Syntax-Instruct} dataset is acceptable. In this study, we utilized this dataset to fine-tune several language models, aiming to enhance their ability to perform the task of automatically generating syntax feedback in essays.

\subsection{From LLMs to Automatic Syntax Feedback Generation}
We developed a system aimed at streamlining the production of Syntax Feedback for essays crafted by students, thereby offering an indispensable resource to educators. This innovation stemmed from a comprehensive process of fine-tuning various Large Language Models (LLMs), including but not limited to the \textit{Llama-2-7b-chat-hf}, \textit{Llama-2-13b-chat-hf} \cite{touvron2023llama}, and the \textit{Mistral-7B-Instruct-v0.2} \cite{jiang2023mistral}. This refinement process leveraged the rich dataset detailed in Section \ref{sec:dataset}, ensuring the feedback generated is both relevant and constructive.\\
By furnishing educators with this system, they are empowered to produce nuanced feedback, which is pivotal in helping students enhance their writing capabilities. Our initiative extends beyond merely developing a tool; it encompasses a thorough examination and assessment of various models. This diligence ensures that the system’s underlying language models are optimally selected and tailored to produce feedback that is not only accurate but also pedagogically valuable. Our work places special emphasis on the meticulous evaluation of the chosen LLMs, aiming to refine and perfect the quality of feedback generated.

\section{Experiments}\label{sec:experiments}
 We explore how the \texttt{Essay-Syntax-Instruct} dataset can be utilized to optimize various families of Large Language Models (LLMs) across multiple dimensions.
 
\subsection{Experimental Setup} 
\paragraph{Data.} We utilized Large Language Models (LLMs) that underwent instruction tuning with \texttt{Essay-Syntax-Instruct} dataset. For evaluation purposes, we designated the 300 annotated instances as a testing set and assessed all our models using GPT-3.5 Turbo as a reference point. The training was conducted with the other 8,020 examples.
\paragraph{Baselines.} 
Our examination considers three Large Language Models (LLMs) specifically calibrated for instruction handling: \textit{LLlama-2-chat-hf}, as documented by \cite{touvron2023llama}, available in both 7B and 13B configurations; and \textit{Mistral-7B-Instruct-v0.2} introduced by \cite{jiang2023mistral}, which is offered in the 7B Parameters.
\paragraph{Training details.}
Fine-tuning of all models was performed using LORA \cite{hu2021lora}, with specifications of $r=32$ and $\alpha=64$ across three training epochs, and a total batch size maintained at 16. The initial learning rate was set to $3\cdot 10^{-4}$, accompanied by a cosine warm-up with a ratio of 0.1. During inference, model distribution sampling facilitated clue generation, setting the temperature at 0.3. Moreover, the top-$p$ and top-$k$ parameters were adjusted to 0.95 and 50, respectively, following the methodology described by Holtzman et al. (2019). This entire experimental setup was executed on a server outfitted with dual NVIDIA A6000 GPUs, leveraging the capabilities of DeepSpeed \cite{rasley2020deepspeed} and FlashAttention 2 \cite{dao2023flashattention}.

\subsection{Results}
\begin{figure*}[ht]
    \centering 
    \includegraphics[width=0.9\textwidth]{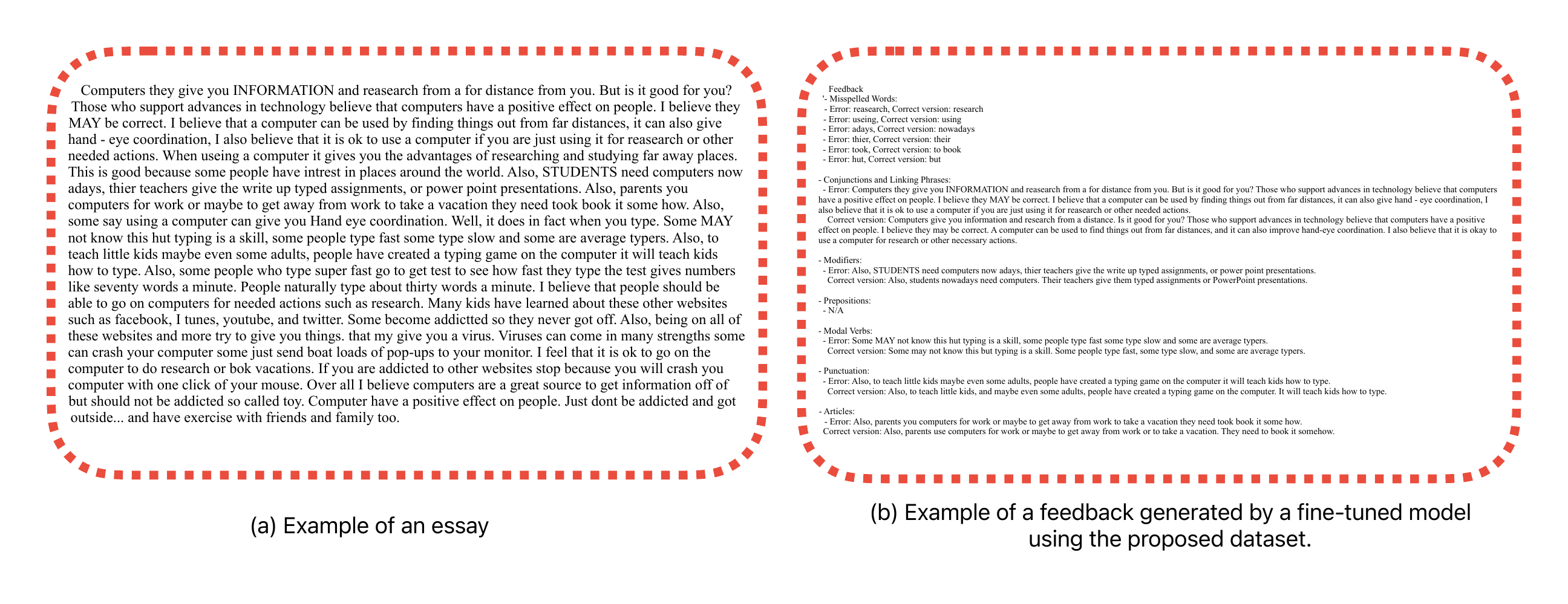}
    \caption{Example of an essay (a) with feedback generated (b) by a fine-tuned model using the proposed dataset.}
    \label{fig:fig7}
\end{figure*}

To evaluate the results, we implemented an evaluation process that entailed generating data using both the base and fine-tuned models on a test dataset comprised of 300 essays, which were not utilized during the training phase. Subsequently, we conducted two distinct types of evaluations. Firstly, we assessed the generated essays utilizing automatic evaluation metrics, specifically Rouge scores. Secondly, we employed human evaluation to assess the generated feedback in Figure \ref{fig:fig7}, you can see an example of an essay along with the feedback generated by a fine-tuned model using the proposed dataset.
Fine-tuning enhances the model's ability to provide precise and contextually appropriate feedback.  We will provide detailed information about this result in the following sections.

\paragraph{Automatic Evaluation}

We attempted to calculate the ROUGE scores between the generated feedback from GPT-3.5-Turbo and other models, which we utilized for fine-tuning in two distinct scenarios: the base model before fine-tuning and the fine-tuned model afterward, using the Essay-Syntax-Instruct dataset. The results of these experiments, presented in Table \ref{tab:tab1}, illustrate that after fine-tuning, there was an increase in the ROUGE scores across all models included in this study. 
Considering that the ROUGE score primarily measures word overlap between the reference and target texts, it may not be highly reliable. This is because the feedback, although potentially insightful, could differ from the assessments provided by  \textit{GPT3.5-Turbo} While ROUGE can offer some useful ideas, its trustworthiness is limited. To address this issue, we applied human evaluation to achieve a more comprehensive understanding of the efficiency of fine-tuning the models using the fine-tuned versions. 
 \begin{figure*}[ht]
    \centering 
    \includegraphics[width=0.9\textwidth]{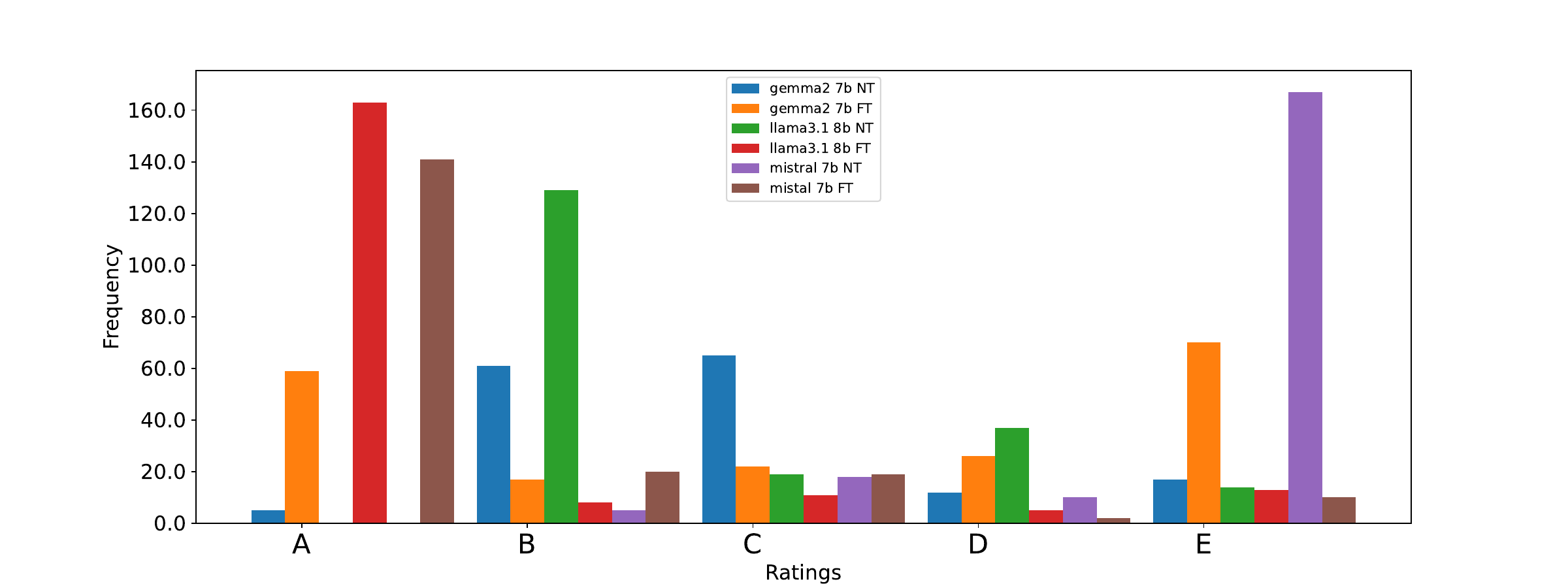}
    \caption{Human evaluation of syntax feedback generated by the LLMs.}
    \label{fig:fig6}
\end{figure*}

\begin{table*}[!ht]
     \centering
     \begin{tabular}{cccccc}
      \hline
    \textbf{model type}&    \textbf{model name} & \textbf{\# params} & \textbf{ROUGE-1}& \textbf{ROUGE-2} & \textbf{ROUGE-L} \\ \hline
     &\textsc{Llama2-chat}  & 7B & 0.351 & 0.146 & 0.185\\
   Base LLMs  &\textsc{Mistral-instruct}  & 7B & \textbf{0.366} & 0.144 & 0.205\\
    &\textsc{Llama2-chat}  & 13B & 0.328 & \textbf{0.148} & \textbf{0.214}\\
     \hline
   
    &\textsc{Llama2-chat}  & 7B & 0.477 & 0.337 & 0.343\\
 
   Finetuned LLMs   &\textsc{Mistral-instruct}  & 7B & \textbf{0.514} & \textbf{0.375} & \textbf{0.369}\\
    &\textsc{Llama2-chat}  & 13B & 0.476 & 0.340 & 0.347\\   \hline
     \end{tabular}
     \caption{Performance of base LLMs and fine-tuned LLMs using the \texttt{Essay-Syntax-Instruct}}
     \label{tab:tab1}
 \end{table*}

 \begin{table*}[!ht]
    \centering
    \begin{tabular}{ccccccc}
        \hline
     & \textbf{Llama2 NF} & \textbf{Llama2 F} & \textbf{Llama2 NF}& \textbf{Llama2 F} & \textbf{Mistral NF} & \textbf{Mistral F}\\
    \hline
    \textbf{\# params}  &7B &7B & 13B& 13B&7B &7B\\
    \hline
    \textbf{Ratings \%}&&&&&&\\
    \hline
    \textbf{A} & 1.00 & 2.67 & 2.00 & 4.33 & 4.00 & 4.67 \\
    \textbf{B} & 24.67 & 44.67 & 11.00 & \textbf{60.00} & 10.33 & \textbf{65.67} \\
    \textbf{C} & 34.33 & \textbf{48.00} & \textbf{56.00} & 24.67 & 22.67 & 22.33 \\
    \textbf{D} & \textbf{39.67} & 1.67 & 24.67 & 5.67 & \textbf{43.33} & 4.00 \\
   \textbf{E} & 0.33 & 3.00 & 6.33 & 5.33 & 19.67 & 3.33 \\
    
    \hline
     \end{tabular}
    \caption{Assessing the percentage of human evaluation for syntax feedback generated by the LLMs for each rating.}
    \label{tab:tab2}
   
\end{table*}

\paragraph{Human Evaluation}

For the qualitative evaluation of generated syntax feedback, we employed two human raters who are master's students from the Department of Linguistics. Their advanced academic background and expertise in linguistics ensure that the feedback evaluation is conducted with a high degree of proficiency and reliability. This human assessment complements our quantitative metrics, providing a holistic evaluation of the feedback quality. Our human annotators reviewed the outputs from the fine-tuned language models (LLMs). We focused this comparison on a subset of 300 essays from the test dataset, employing the same guidelines introduced earlier for validating the generated dataset, as discussed in a previous section. The findings of this evaluation are detailed in Figure~\ref{fig:fig6} and Table \ref{tab:tab2}.\\
Our analysis revealed that fine-tuning significantly enhances the performance of large language models in generating syntax feedback for essays. The fine-tuned models demonstrate substantial improvements in achieving higher ratings (A and B), coupled with a notable reduction in poorer ratings (D). This trend is consistent across different model configurations and sizes, affirming that fine-tuning effectively optimizes models for specific tasks, resulting in more accurate and useful feedback in educational contexts.We observed that language models initially produced several errors in syntax feedback, including repeated mistakes across different categories and incorrect highlighting of correct words. However, fine-tuning significantly mitigates these errors, leading to more accurate and constructive feedback.\\
\paragraph*{Llama2 (7 Billion Parameters)}
Fine-tuning significantly enhanced the capability of the model, resulting in a nearly tripled score for the best feedback (Rating A) from 1.00\% to 2.67\%. There was also a marked increase in good quality feedback (Rating B) from 24.67\% to 44.67\% and above-average feedback (Rating C) from 34.33\% to 48.00\%. Concurrently, there was a dramatic reduction in poorer quality feedback (Rating D) from 39.67\% to 1.67\%, demonstrating a substantial improvement in feedback quality. The lowest feedback category (Rating E) had only a minimal increase from 0.33\% to 3.00\%.

\paragraph*{Llama2 (13 Billion Parameters)}
This version showed profound improvements in delivering the best feedback (Rating A), improving from 2.00\% to 4.33\%, and good feedback (Rating B) from 11.00\% to 60.00\%. There was a substantial shift towards higher feedback quality with a significant reduction in average feedback (Rating C) from 56.00\% to 24.67\% and poor feedback (Rating D) from 24.67\% to 5.67\%. The worst feedback category (Rating E) slightly declined from 6.33\% to 5.33\%, indicating improved consistency in feedback quality.

\paragraph*{Mistral (7 Billion Parameters)}
Post-fine-tuning, this model slightly improved top-quality feedback (Rating A) from 4.00\% 
These outcomes indicate that fine-tuning improves the alignment of generated content with the anticipated format. Moreover, it also positively influences the overall quality of the generated content, suggesting that fine-tuning not only corrects formatting issues but also enhances the essence of the content.

\section{Conclusion}\label{sec:conclusions}

In this research, we have developed an innovative tool, the Automatic Essay Syntax Feedback Generator, which leverages the capabilities of Large Language Models (LLMs) to provide syntax feedback for educational purposes. This cutting-edge system is designed to boost student interaction and enhance the retention of learning material through dynamic engagement.\\
Furthermore, our study contributes significantly to the expansion of essay syntax feedback resources by introducing \texttt{Essay-Syntax-Instruct} dataset. We have compiled a thorough dataset comprising essays written by students alongside the syntax feedback provided. This dataset is indispensable for the development of educational systems and analytical instruments, paving the way for new research and innovation opportunities.\\
Looking forward, we are dedicated to broadening the scope of our method to encompass more languages, thereby increasing the accessibility of this educational tool. We also aim to refine our syntax feedback methodologies by incorporating more sophisticated LLMs, thereby pushing the boundaries of what's possible in educational technology. In future work, we will conduct real-world case studies to assess the practical impact of automated syntax feedback on student learning outcomes. This will help us determine our findings' applicability to various educational settings. Further research will focus on comparing our system with traditional teaching methods to evaluate its effectiveness. Additionally, We aim to streamline the fine-tuning process of LLMs, making it more efficient and accessible for educational institutions with limited computational resources and expertise.

\bibliographystyle{plainnat}

\begin{thebibliography}{8}
\bibitem{ramesh2022automated}Ramesh, D. \& Sanampudi, S. An automated essay scoring systems: a systematic literature review. {\em Artificial Intelligence Review}. \textbf{55}, 2495-2527 (2022)
\bibitem{miltsakaki2004evaluation}Miltsakaki, E. \& Kukich, K. Evaluation of text coherence for electronic essay scoring systems. {\em Natural Language Engineering}. \textbf{10}, 25-55 (2004)
\bibitem{ridley2020prompt}Ridley, R., He, L., Dai, X., Huang, S. \& Chen, J. Prompt agnostic essay scorer: a domain generalization approach to cross-prompt automated essay scoring. {\em ArXiv Preprint ArXiv:2008.01441}. (2020)
\bibitem{mikolov2013efficient}Mikolov, T., Chen, K., Corrado, G. \& Dean, J. Efficient estimation of word representations in vector space. {\em ArXiv Preprint ArXiv:1301.3781}. (2013)
\bibitem{pennington2014glove}Pennington, J., Socher, R. \& Manning, C. Glove: Global vectors for word representation. {\em Proceedings Of The 2014 Conference On Empirical Methods In Natural Language Processing (EMNLP)}. pp. 1532-1543 (2014)
\bibitem{sultan2016fast}Sultan, M., Salazar, C. \& Sumner, T. Fast and easy short answer grading with high accuracy. {\em Proceedings Of The 2016 Conference Of The North American Chapter Of The Association For Computational Linguistics: Human Language Technologies}. pp. 1070-1075 (2016)
\bibitem{mathias2018thank}Mathias, S. \& Bhattacharyya, P. Thank “goodness”! a way to measure style in student essays. {\em Proceedings Of The 5th Workshop On Natural Language Processing Techniques For Educational Applications}. pp. 35-41 (2018)
\bibitem{salim2019automated}Salim, Y., Stevanus, V., Barlian, E., Sari, A. \& Suhartono, D. Automated English digital essay grader using machine learning. {\em 2019 IEEE International Conference On Engineering, Technology And Education (TALE)}. pp. 1-6 (2019)
\bibitem{dong2016automatic}Dong, F. \& Zhang, Y. Automatic features for essay scoring–an empirical study. {\em Proceedings Of The 2016 Conference On Empirical Methods In Natural Language Processing}. pp. 1072-1077 (2016)
\bibitem{taghipour2016neural}Taghipour, K. \& Ng, H. A neural approach to automated essay scoring. {\em Proceedings Of The 2016 Conference On Empirical Methods In Natural Language Processing}. pp. 1882-1891 (2016)
\bibitem{riordan2017investigating}Riordan, B., Horbach, A., Cahill, A., Zesch, T. \& Lee, C. Investigating neural architectures for short answer scoring. {\em Proceedings Of The 12th Workshop On Innovative Use Of NLP For Building Educational Applications}. pp. 159-168 (2017)
\bibitem{rodriguez2019language}Rodriguez, P., Jafari, A. \& Ormerod, C. Language models and automated essay scoring. {\em ArXiv Preprint ArXiv:1909.09482}. (2019)
\bibitem{lun2020multiple}Lun, J., Zhu, J., Tang, Y. \& Yang, M. Multiple data augmentation strategies for improving performance on automatic short answer scoring. {\em Proceedings Of The AAAI Conference On Artificial Intelligence}. \textbf{34}, 13389-13396 (2020)
\bibitem{devlin2018bert}Devlin, J., Chang, M., Lee, K. \& Toutanova, K. Bert: Pre-training of deep bidirectional transformers for language understanding. {\em ArXiv Preprint ArXiv:1810.04805}. (2018)
\bibitem{yang2020enhancing}Yang, R., Cao, J., Wen, Z., Wu, Y. \& He, X. Enhancing automated essay scoring performance via fine-tuning pre-trained language models with combination of regression and ranking. {\em Findings Of The Association For Computational Linguistics: EMNLP 2020}. pp. 1560-1569 (2020)
\bibitem{wang2022use}Wang, Y., Wang, C., Li, R. \& Lin, H. On the use of bert for automated essay scoring: Joint learning of multi-scale essay representation. {\em ArXiv Preprint ArXiv:2205.03835}. (2022)
\bibitem{mizumoto2023exploring}Mizumoto, A. \& Eguchi, M. Exploring the potential of using an AI language model for automated essay scoring. {\em Research Methods In Applied Linguistics}. \textbf{2}, 100050 (2023)
\bibitem{naismith2023automated}Naismith, B., Mulcaire, P. \& Burstein, J. Automated evaluation of written discourse coherence using GPT-4. {\em Proceedings Of The 18th Workshop On Innovative Use Of NLP For Building Educational Applications (BEA 2023)}. pp. 394-403 (2023)
\bibitem{yancey2023rating}Yancey, K., Laflair, G., Verardi, A. \& Burstein, J. Rating short l2 essays on the cefr scale with gpt-4. {\em Proceedings Of The 18th Workshop On Innovative Use Of NLP For Building Educational Applications (BEA 2023)}. pp. 576-584 (2023)
\bibitem{yan2024practical}Yan, L., Sha, L., Zhao, L., Li, Y., Martinez-Maldonado, R., Chen, G., Li, X., Jin, Y. \& Gašević, D. Practical and ethical challenges of large language models in education: A systematic scoping review. {\em British Journal Of Educational Technology}. \textbf{55}, 90-112 (2024)
\bibitem{kasneci2023chatgpt}Kasneci, E., Seßler, K., Küchemann, S., Bannert, M., Dementieva, D., Fischer, F., Gasser, U., Groh, G., Günnemann, S., Hüllermeier, E. \& Others ChatGPT for good? On opportunities and challenges of large language models for education. {\em Learning And Individual Differences}. \textbf{103} pp. 102274 (2023)
\bibitem{peng2023check}Peng, B., Galley, M., He, P., Cheng, H., Xie, Y., Hu, Y., Huang, Q., Liden, L., Yu, Z., Chen, W. \& Others Check your facts and try again: Improving large language models with external knowledge and automated feedback. {\em ArXiv Preprint ArXiv:2302.12813}. (2023)
\bibitem{han2023fabric}Han, J., Yoo, H., Myung, J., Kim, M., Lim, H., Kim, Y., Lee, T., Hong, H., Kim, J., Ahn, S. \& Others Fabric: Automated scoring and feedback generation for essays. {\em ArXiv Preprint ArXiv:2310.05191}. (2023)
\bibitem{wang2022self}Wang, Y., Kordi, Y., Mishra, S., Liu, A., Smith, N., Khashabi, D. \& Hajishirzi, H. Self-instruct: Aligning language models with self-generated instructions. {\em ArXiv Preprint ArXiv:2212.10560}. (2022)
\bibitem{jiang2023mistral}Jiang, A., Sablayrolles, A., Mensch, A., Bamford, C., Chaplot, D., Casas, D., Bressand, F., Lengyel, G., Lample, G., Saulnier, L. \& Others Mistral 7B. {\em ArXiv Preprint ArXiv:2310.06825}. (2023)
\bibitem{rasley2020deepspeed}Rasley, J., Rajbhandari, S., Ruwase, O. \& He, Y. Deepspeed: System optimizations enable training deep learning models with over 100 billion parameters. {\em Proceedings Of The 26th ACM SIGKDD International Conference On Knowledge Discovery \& Data Mining}. pp. 3505-3506 (2020)
\bibitem{dao2023flashattention}Dao, T. Flashattention-2: Faster attention with better parallelism and work partitioning. {\em ArXiv Preprint ArXiv:2307.08691}. (2023)
\bibitem{hu2021lora}Hu, E., Shen, Y., Wallis, P., Allen-Zhu, Z., Li, Y., Wang, S., Wang, L. \& Chen, W. Lora: Low-rank adaptation of large language models. {\em ArXiv Preprint ArXiv:2106.09685}. (2021)
\bibitem{xiao2024automation}Xiao, C., Ma, W., Xu, S., Zhang, K., Wang, Y. \& Fu, Q. From Automation to Augmentation: Large Language Models Elevating Essay Scoring Landscape. {\em ArXiv Preprint ArXiv:2401.06431}. (2024)
\bibitem{touvron2023llama}Touvron, H., Martin, L., Stone, K., Albert, P., Almahairi, A., Babaei, Y., Bashlykov, N., Batra, S., Bhargava, P., Bhosale, S. \& Others Llama 2: Open foundation and fine-tuned chat models. {\em ArXiv Preprint ArXiv:2307.09288}. (2023)

\end{thebibliography}

\end{document}